# A HYBRID LEARNING ALGORITHM FOR TEXT CLASSIFICATION

*S. M. Kamruzzaman and Farhana Haider*
Department of Computer Science and Engineering
International Islamic University Chittagong, Chittagong-4203, Bangladesh
Email: smk_iiuc@yahoo.com, farhanahdr@yahoo.com

## ABSTRACT

Text classification is the process of classifying documents into predefined categories based on their content. Existing supervised learning algorithms to automatically classify text need sufficient documents to learn accurately. This paper presents a new algorithm for text classification that requires fewer documents for training. Instead of using words, word relation i.e association rules from these words is used to derive feature set from pre-classified text documents. The concept of Naïve Bayes classifier is then used on derived features and finally only a single concept of Genetic Algorithm has been added for final classification. Experimental results show that the classifier build this way is more accurate than the existing text classification systems.

## 1. INTRODUCTION

Text classification has become one of the most important techniques in the text data mining. The task is to automatically classify documents into predefined classes based on their content. Many algorithms have been developed to deal with automatic text classification [3]. With the existing algorithms, a number of newly established processes are involving in the automation of text classification. It has been observed that for the purpose of text classification the concept of association rule is very well known. Association rule mining [1] finds interesting association or correlation relationships among a large set of data items [4]. The discovery of these relationships among huge amounts of transaction records can help in many decision making process. On the other hand, the Naïve Bayes classifier uses the maximum a posteriori estimation for learning a classifier. It assumes that the occurrence of each word in a document is conditionally independent of all other words in that document given its class [3]. Although the Naïve Bayes works well in many studies [7], it requires a large number of training documents for learning accurately. Genetic algorithm starts with an initial population which is created consisting of randomly generated rules. Each rule can be represented by a string of bits. Based on the notion of survival of the fittest, a new population is formed to consist of the fittest rules in the current population, as well as offspring of these rules. Typically, the fitness of a rule is assessed by its classification accuracy on a set of training examples.

This paper presents a new algorithm for text classification. Instead of using words, word relation i.e association rules is used to derive feature set from pre-classified text documents. The concept of Naive Bayes Classifier is then used on derived features and finally a concept of Genetic Algorithm has been added for final classification. A system based on the proposed algorithm has been implemented and tested. The experimental results show that the proposed system works as a successful text classifier.

## 2. PREVIOUS WORK

### 2.1 Classification using Association Rule with Naïve Bayes Classifier

Text classification using the concept of association rule of data mining where Naïve Bayes classifier was used to classify text finally, showed the dependability of the Naïve Bayes classifier with association rules [4]. But since this method ignores the concept of calculation of negative example for any specific class determination, the accuracy may fall in some cases. For classifying a text it just calculates the probability of different class with the probability values of the matched set while ignoring the unmatched sets.

### 2.2 Classification using Association Rule with Decision Tree

Text classification using decision tree showed an acceptable accuracy using 76% training data of total data set [8], while it is possible to achieve good accuracy using only 40 to 50% of total data sets as training data.

### 2.3 Work on Genetic Algorithm

Text Classification based on genetic algorithm showed satisfactory performance using 69% training data, but this process requires the time consuming steps to classify the texts [2][5].

## 3. THE PROPOSED METHOD

The concept of positive and negative examples has been observed in the following algorithm. The associated word sets, which do not mach our considered class is, treated as negative sets and others are positive.

### 3.1 Proposed Algorithm

n = number of class, m = number of associated sets
1. For each class i = 1 to n do
2. Set pval = 0, nval = 0, p = 0, n = 0
3. For each set s = 1 to m do
4. If the probability of the class (i) for the set (s) is maximum then increment pval else increment nval
5. If 50% of the associated set s is matched with the keywords set do step 6 else do step 7
6. If maximum probability matches the class i then increment p
7. If maximum probability does not match the class i increment n
8. If (s<=m) go to step 3
9. Calculate the percentage of matching in positive sets for the class i
10. Calculate the percentage of not matching in negative sets for the class i
11. Calculate the total probability as the summation of the results obtained from step 9 and 10 and also the prior probability of the class i in set s
12. If (i<=n) go to step 1
13. Set the class having the maximum probability value as the result

### 3.2 Preparing Text for Classification

We used abstracts of different research papers some of them are from the proceedings of ICCIT 2002 and the rest of them are from World Wide Web. Three classes of papers from Algorithm (ALG), Educational Engineering (EDE) and Artificial Intelligence (AI) are considered for our experiment. Total 103 abstracts (27 from Algorithm, 14 from Educational Engineering and 62 from AI) have been used for the experiment. To clean the text we have considered only the keywords. That is unnecessary words and symbols are removed. For this keyword extraction process we dropped the common unnecessary words like am, is, are, to, from...etc. and also dropped all kinds of punctuations and stop words. Singular and plural form of a word is considered same. Finally, the remaining frequent words are considered as keywords. Let an abstract:

*Let G = (V, E) be a connected graph, and let X be a vertex subset of G. Let f be a mapping from X to the set of natural numbers such that f(x) ³ 2 for all x in X. A degree restricted spanning tree is a spanning tree T of G such that f(x) £ degT(x) for all x belongs X, where degT(x) denotes the degree of a vertex x in T. In this paper, we show that the decision problem "whether there exists a degree restricted spanning tree in G" is NP-complete. We also give a restricted proof of a conjecture, provided by Kaneko and Yoshimoto, on the existence of such a spanning tree in general graphs. Finally, we present a polynomial-time algorithm to find a degree restricted spanning tree of a graph satisfying the conditions presented in the restricted proof of the conjecture.*

Keywords extracted from this abstract are: Spanning, tree, bipartite, graph. This keyword extraction process is applied to all the abstracts.

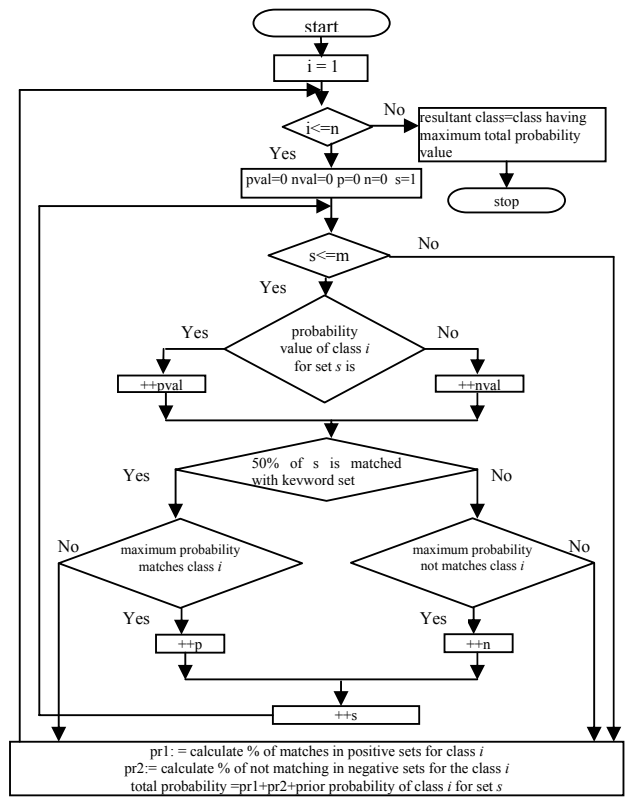

**Fig. 1** Flow chart of the proposed algorithm

## 4. EXPERIMENTAL RESULTS

### 4.1 Training Phase

#### 4.1.1 Deriving Associated Word Sets

Each abstract is considered as a transaction in the text data. After pre-processing the text data association rule mining [1] is applied to the set of transaction data where each frequent word set from each abstract is considered as a single transaction. Using these transactions, we generated a list of maximum length sets applying the Apriori algorithm [1]. The support and confidence is set to 0.05 and 0.75 respectively. A partial list of the generated large word set with their occurrence frequency is illustrated in Table 1.

**Table 1** Word set with occurrence frequency

| Maximum Length Set | No of Occurrence | | |
|---|---|---|---|
| | ALG | EDE | AI |
| neural, network | | | 5 |
| gray, code | 4 | | |
| set, length | 2 | | |
| expectation, teacher, found, significant | | 2 | |
| education, level, test, significant, difference | | 2 | |
| significant, study, result | | 3 | |
| Lossless, compression | | | 2 |

### 4.1.2 Setting Associated Word Set With Probability Value

There exist some algorithms for learning to classify text based on the Naïve Bayes classifier. In applying Naïve Bayes classifier, each word position in a document is defined as an attribute and the value of that attribute to be the English word found in that position. Naïve Bayes classification is given by:

$$V_{NB} = \text{argmax}\ P(V_j) \prod P(a_i | V_j)$$

To summarize, the Naïve Bayes classification $V_{NB}$ is the classification that maximizes the probability of observing the words that were actually found in the example documents, subject to the usual Naïve Bayes independence assumption. The first term can be estimated based on the fraction of each class in the training data. The following equation is used for estimating the second term:

$$\frac{n_k + 1}{n + |\text{vocabulary}|} \quad \quad (1)$$

where $n$ is the total number of word positions in all training examples whose target value is $V_j$, $n_k$ is the number of times that word is found among these $n$ words positions, and $|\text{vocabulary}|$ is the total number of distinct words found within the training data.

From the generated word set after applying association mining on training data we have found the following information: total number of word set is 20, total number of word set from Algorithm (ALG), Educational Engineering (EDE) and Artificial Intelligence (AI) are 6, 7, 7 respectively. Now we can use the Naïve Bayes classifier for probability calculation. The calculation of first term is based on the fraction of each target class in the training data. Prior probability for ALG, EDE and AI are 0.3, 0.35 and 0.35 respectively. Then the second term is calculated according to the equation (1). The probability values of word set are listed in Table 2.

**Table 2** Word set with probability value

| Maximum Length Set | Probability | | |
|---|---|---|---|
| | ALG | EDE | AI |
| neural, network | 0.024 | 0.024 | 0.146 |
| gray, code | 0.217 | 0.043 | 0.043 |
| set, length | 0.130 | 0.043 | 0.043 |
| expectation, teacher, found, significant | 0.058 | 0.176 | 0.058 |
| education, level, test, significant, difference | 0.058 | 0.176 | 0.058 |
| significant, study, result | 0.058 | 0.235 | 0.058 |
| Lossless, compression | 0.024 | 0.024 | 0.073 |

### 4.2 Testing Phase

#### 4.2.1 Classifying a New Document

As for example let us consider the extracted sets of keywords of a new abstract are: study, framework, evaluation, open, learning, environment, methodology, role, field, using, student, task, regarding, issue, design.
For this set of keywords,
Calculated Probability for class ALG = 88.913
Calculated Probability for class EDE = 116.691
Calculated Probability for class AI = 87.573
*From the above result we found the abstract belongs to class Educational Engineering (EDE).*

## 5. COMPARATIVE STUDY

### 5.1 Association Rule and Naïve Bayes Classifier

The following results are found using the same data sets for both Association Rule with Naive Bayes Classifier and proposed method. The result shows that proposed approach work well using only 50% Training data.

### 5.2 Association Rule Based Decision Tree

In text categorization using association rule based decision tree 76 % data set of the total 33 data set was used to train and 13% error observed. On the other hand using only 50% data as training the proposed algorithm can able to classify text with 78% accuracy rate. The major problem of decision tree based classified is that, this system is totally failed to categorize a class. Our proposed technique shows better performance even with 3 times larger data sets.

**Table 3** Comparison of proposed method with text classifier using association rule and naïve Bayes classifier

| % of Training Data | % of Accuracy | |
|---|---|---|
| | Association Rule with Naïve Byes Classifier | Proposed Method |
| 10 | 36 | 13 |
| 20 | 40 | 76 |
| 30 | 63 | 85 |
| 40 | 63 | 71 |
| 50 | 31 | 78 |

Accuracy Vs Training Data

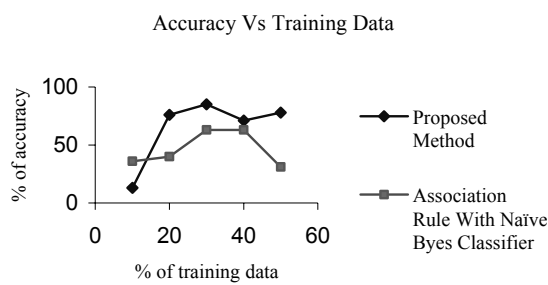

**Fig. 2** Accuracy Vs Training Curve

### 5.3 Genetic Algorithm

Researchers showed 68% accuracy using the concept of genetic algorithm with 31% test data [5] while our technique is better both in accuracy and % of test data. Moreover it required processing for each class during training. But our proposed algorithm does not require such process during training phase and hence reduces time.

**Table 4** Comparison of proposed method with text classifier using decision tree and genetic algorithm

| Technique | (%) Training Data | (%) Accuracy |
|---|---|---|
| Association Rule Based Decision Tree | 76 | 87 |
| Genetic Algorithm | 69 | 68 |
| Proposed Algorithm | 50 | 78 |

### 6. FUTURE WORK

If we form the Frequent Pattern (FP) growth tree, time would be shorter enough. Even the experimental results are quite encouraging, it would be better if we work with larger data sets. More features from Genetic Algorithm can be made the technique to classify text more efficiently. In near future, we will try to apply this algorithm for emerging pattern identification [6].

### 7. CONCLUSION

This paper presented a new hybrid technique for text classification. The existing algorithms require more data for training as well as the computational time of these algorithms also increases. In contrast to the existing algorithms, the proposed hybrid algorithm requires less training data and less computational time. In spite of the randomly chosen training set we achieved 78% accuracy for 50% training data. Though 85% accuracy was observed in 30% training data, a class could not be classified, so we dropped this position and increased training data set for more acceptable result.